%% file: main.tex
\documentclass[10pt,twocolumn,letterpaper]{article}

\usepackage[pagenumbers]{cvpr} 

\input{preamble}

\definecolor{cvprblue}{rgb}{0.21,0.49,0.74}
\usepackage[pagebackref,breaklinks,colorlinks,citecolor=cvprblue]{hyperref}

\usepackage{graphicx}
\usepackage{subcaption}  
\usepackage{pifont}
\usepackage{algorithm}  
\usepackage{algorithmic} 
\usepackage{amsmath}     
\usepackage{multicol}
\usepackage{multirow}
\usepackage[accsupp]{axessibility} 

\def\paperID{5673} 
\def\confName{CVPR}
\def\confYear{2024}

\title{CDFormer: When Degradation Prediction Embraces Diffusion Model for Blind Image Super-Resolution}

\author{Qingguo Liu, ~Chenyi Zhuang, ~Pan Gao\thanks{Corresponding authors.}, ~Jie Qin\footnotemark[1]\\
College of Computer Science and Technology, Nanjing University of Aeronautics and Astronautics \\
{\tt\small \{liuqingguo, chenyi.zhuang, pan.gao, jie.qin\}@nuaa.edu.cn}
}

\begin{document}
\maketitle
\input{sec/0_abstract}    
\input{sec/1_intro}
\input{sec/2_relwork}
\input{sec/3_method}
\input{sec/4_experiments}
\input{sec/5_conclusion}

{
    \small
    \bibliographystyle{ieeenat_fullname}
    \bibliography{main}
}


\end{document}

%% file: preamble.tex
\usepackage[dvipsnames]{xcolor}
\newcommand{\red}[1]{{\color{red}#1}}


%% file: sec/0_abstract.tex
\begin{abstract}
Existing Blind image Super-Resolution (BSR) methods focus on estimating either kernel or degradation information, but have long overlooked the essential content details. In this paper, we propose a novel BSR approach, Content-aware Degradation-driven Transformer (CDFormer), to capture both degradation and content representations. However, low-resolution images cannot provide enough content details, and thus we introduce a diffusion-based module $CDFormer_{diff}$ to first learn Content Degradation Prior (CDP) in both low- and high-resolution images, and then approximate the real distribution given only low-resolution information. Moreover, we apply an adaptive SR network $CDFormer_{SR}$ that effectively utilizes CDP to refine features. Compared to previous diffusion-based SR methods, we treat the diffusion model as an estimator that can overcome the limitations of expensive sampling time and excessive diversity. Experiments show that CDFormer can outperform existing methods, establishing a new state-of-the-art performance on various benchmarks under blind settings. Codes and models will be available at \href{https://github.com/I2-Multimedia-Lab/CDFormer}{https://github.com/I2-Multimedia-Lab/CDFormer}.
\end{abstract}

%% file: sec/1_intro.tex
\section{Introduction}
\label{sec:intro}
\begin{figure}[htbp]
	\centering
	\begin{subfigure}[b]{0.15\textwidth}
		\centering
		\includegraphics[width=1.0\textwidth]{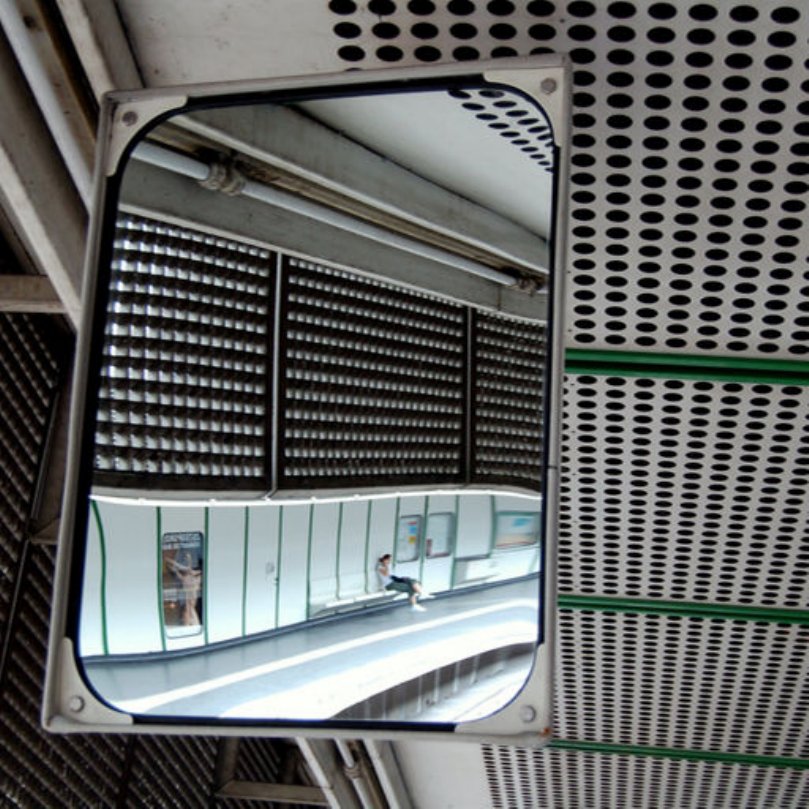}
		\subcaption*{\makebox[9em][c]{HR} \\ \makebox[9em][c]{PSNR$\uparrow$/SSIM$\uparrow$}}
	\end{subfigure}
	\begin{subfigure}[b]{0.15\textwidth}
		\centering
		\includegraphics[width=1.0\textwidth]{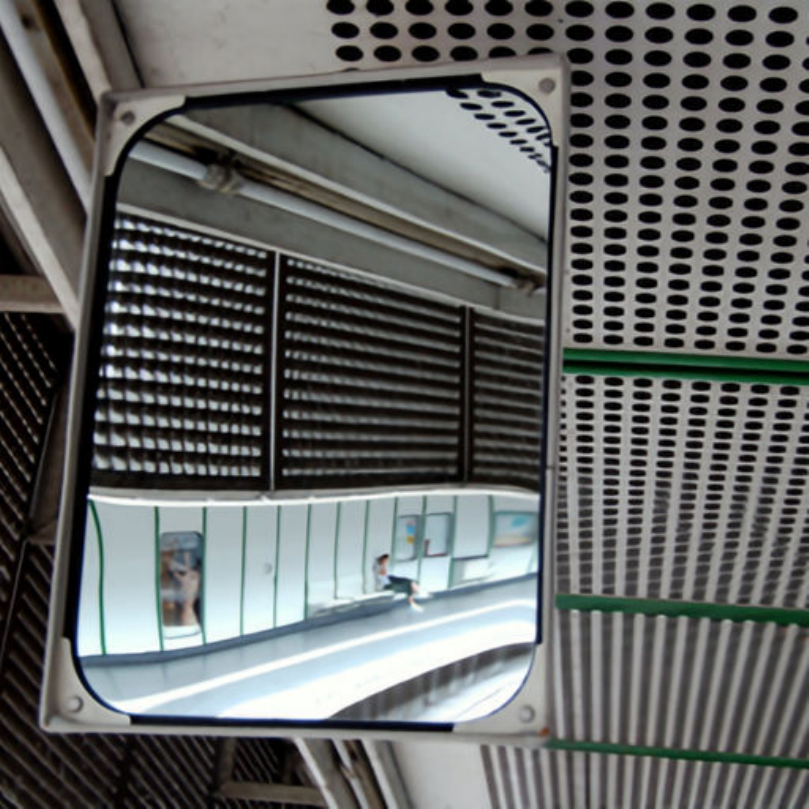}
		\subcaption*{\makebox[9em][c]{DASR\cite{wang2021unsupervised}} \\ \makebox[9em][c]{19.22/0.745}}
	\end{subfigure}
	\begin{subfigure}[b]{0.15\textwidth}
		\centering
		\includegraphics[width=1.0\textwidth]{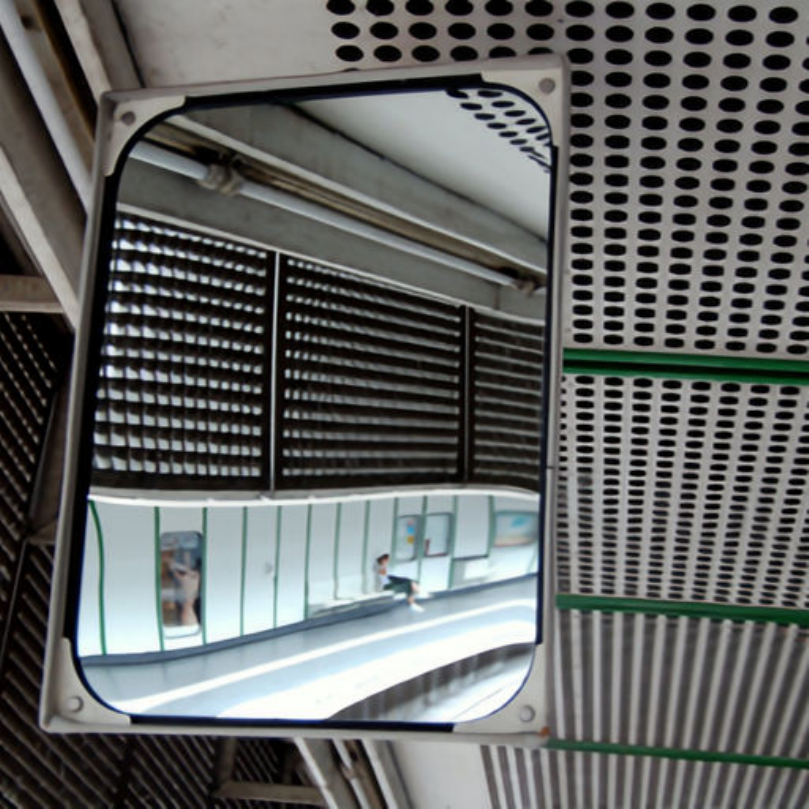}
		\subcaption*{\makebox[9em][c]{KDSR\cite{xia2022knowledge}} \\ \makebox[9em][c]{19.90/0.778}}
	\end{subfigure}

	\begin{subfigure}[b]{0.15\textwidth}
		\centering
		\includegraphics[width=1.0\textwidth]{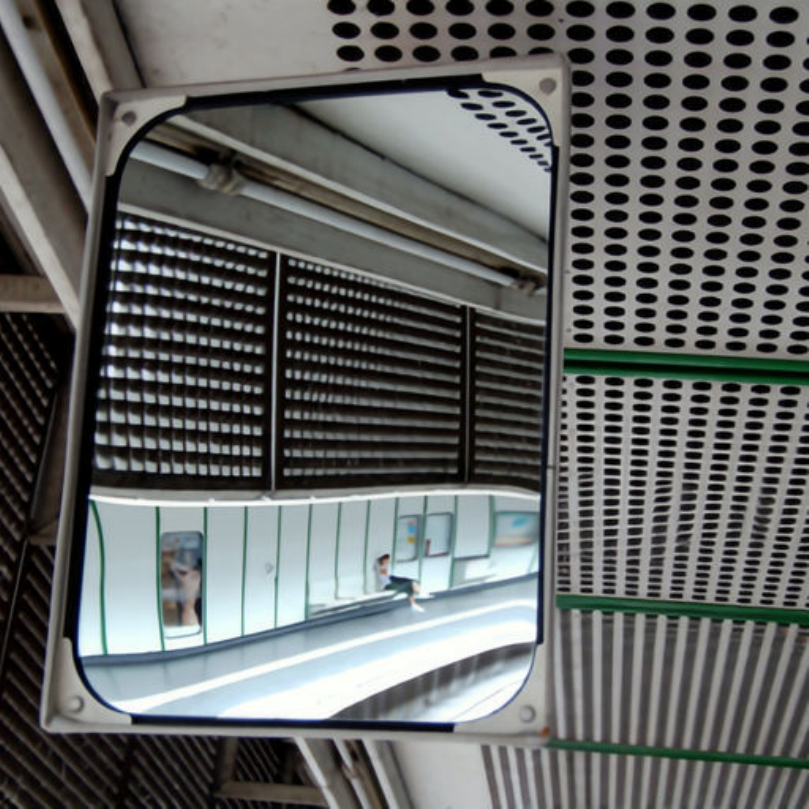}
		\subcaption*{\makebox[9em][c]{DCLS\cite{9879171}} \\ \makebox[9em][c]{21.34/0.851}}
	\end{subfigure}
	\begin{subfigure}[b]{0.15\textwidth}
		\centering
		\includegraphics[width=1.0\textwidth]{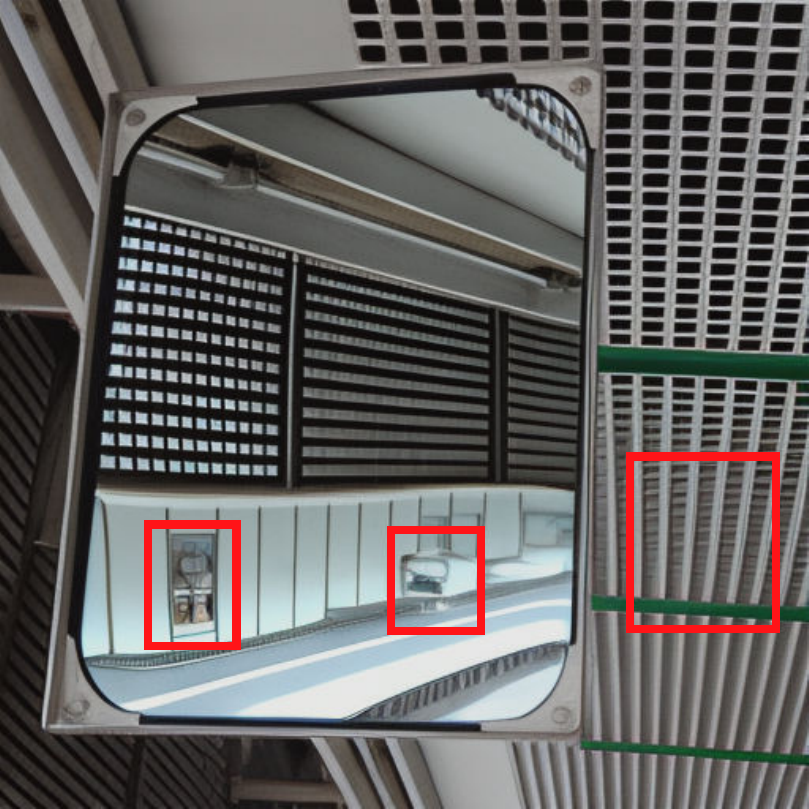}
		\subcaption*{\makebox[9em][c]{StableSR\cite{wang2023exploiting}} \\ \makebox[9em][c]{13.03/0.278}}
	\end{subfigure}
	\begin{subfigure}[b]{0.15\textwidth}
		\centering
		\includegraphics[width=1.0\textwidth]{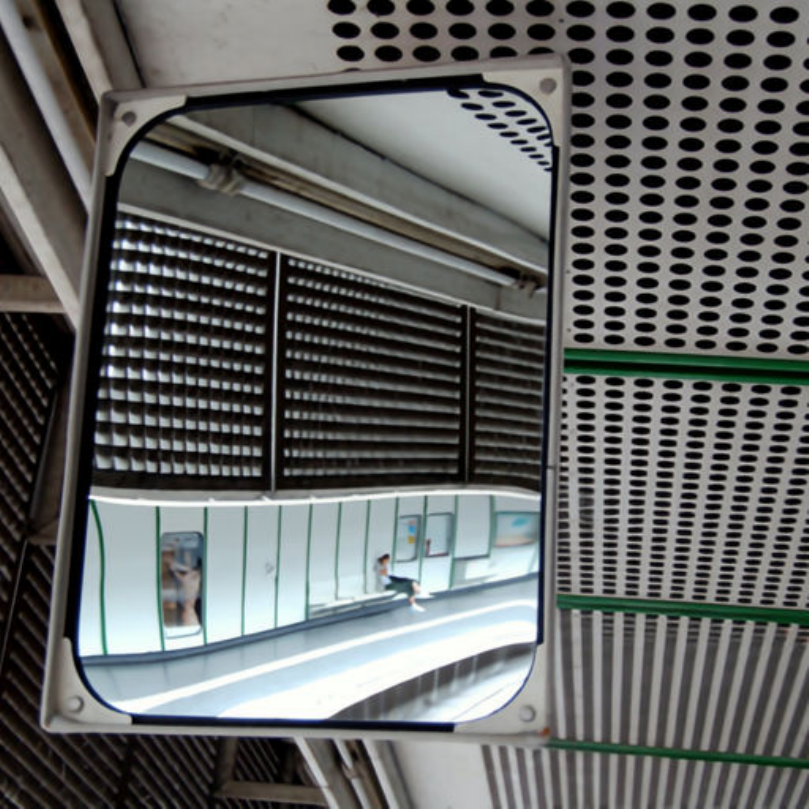}
		\subcaption*{\makebox[9em][c]{CDFormer(Ours)} \\ \makebox[9em][c]{22.66/0.878}}
	\end{subfigure}	
	
	\caption{Blind image Super-Resolution for scale 4 on kernel width 1.2. Our proposed CDFormer with CDP is capable of producing sharp and clean textures and outperforms previous state-of-the-art approaches DASR, KDSR, DCLS, and StableSR.}
    \vspace{-1em}
	\label{fig:4}
\end{figure}
Image Super-Resolution (SR), which aims to reconstruct high-resolution (HR) images from their low-resolution (LR) counterparts, is a long-standing low-level vision problem in the research community. Advanced methods based on deep learning networks start from the assumption that the LR image is degraded from the HR image by a specific process, which can be formulated as follows:
\begin{equation}
\label{eq:1}
I_{LR} = (I_{HR} \ast k_h){\downarrow_s} + n.
\end{equation}
where $k_h$ is a blur kernel, $\downarrow_s$ is a downsampling operation with scale factor $s$, $n$ is additive Gaussian noise. 
In practice, however, the transformation from HR image to LR image is complicated and usually difficult to formulate simply as \cref{eq:1}. As a result, previous works \cite{7115171, 8954252, 8954253, dong2016accelerating, 9578528, 9857155, liang2021swinir, lim2017enhanced, li2021efficient} explore non-blind image Super-Resolution, where the kernel and downsampling are assumed to be known as the prior, have faced great challenges in real-world scenarios, thus extending to Blind image Super-Resolution (BSR). 

Existing BSR methods can be divided into two categories: Kernel Prediction (KP) \cite{liang2021mutual,9879171,bell2019blind,gu2019blind} and Degradation Prediction (DP) \cite{DSAT, wang2021unsupervised, xia2022knowledge}. KP methods produce desirable SR results in most cases, but still face severe performance drops when it has to deal with complex degradations. This inevitable limitation is subject to the unavailability of real-word kernels as well as training distribution mismatch. Furthermore, KP methods have to be limited to blur kernel-based degradation and cannot address other types of degradation (\eg, noise). In contrast, DP methods that DASR \cite{wang2021unsupervised} exemplify, explore degradation representations in an unsupervised manner. However, although DASR can outperform some KP methods \cite{bell2019blind,gu2019blind} for simple degradation, a gap remains between DP and state-of-the-art KP methods \cite{liang2021mutual,9879171} for more complex cases.

More recently, the advances in Diffusion Models (DMs) \cite{sohl2015deep, ho2020denoising, rombach2022high} have revolutionized the community of image synthesis \cite{dhariwal2021diffusion, ho2022cascaded, song2020score} as well as image restoration tasks, such as inpainting \cite{9880056,9879691} and Super-Resolution \cite{9887996, kawar2022denoising}. 
Several works \cite{9887996, li2022srdiff, luo2023refusion, wang2023exploiting} have explored the powerful generative ability of DMs for Super-Resolution and attracted extensive attention. We argue that although these works provide novel viewpoints and solutions for image SR, they suffer from two main drawbacks: (1) treating DMs as the main SR net requires numerous inference steps ($\sim$ $50$ to $1000$ steps), which will be computationally expensive and not suitable for real-time applications. Although the former issue can be alleviated by reducing iteration number \cite{wang2022zero}, it will lead to quality deterioration of the SR results. (2) undesirable artifacts such as joint misalignment or texture distortion have been introduced as a result of the error propagation inherent in single-step noise prediction models.

As shown in \cref{fig:4}, DCLS \cite{9879171}, a state-of-the-art KP method, exhibits significant improvements compared to DP methods, DASR \cite{wang2021unsupervised} and KDSR \cite{xia2022knowledge}, in terms of PSNR and SSIM. Despite being on the leading edge of DM-based SR approaches, StableSR \cite{wang2023exploiting} suffers from incorrect textures (an ellipse is reconstructed as a square) and loss of detail (the person in the mirror disappears). We suspect that the erroneous texture results via StableSR may be blamed on the diversity in the pre-trained DMs which is excessive for SR task. Due to the low quality of given image, the priors in pre-trained DMs may misinterpret the given LR images and reconstruct them in a wrong context.

In this paper, we rethink the existing DP methods that concentrate on estimating degradation representations from LR images while neglecting the essential content information. We propose CDFormer, a Content-aware Degradation-driven Transformer network for BSR. Instead of employing the DM as the whole SR network, we design a Content Degradation Prior (CDP) generation module $CDFormer_{diff}$, where the DM is treated as an estimator to recreate CDP from LR images.
The Content-aware Degradation-driven Transformer SR module $CDFormer_{SR}$ further utilizes the estimated CDP via several injection modules to adaptively refine features and benefit reconstruction.
Our main contributions are as follows:
\begin{itemize}
    \item We introduce a Content Degradation Prior (CDP) generation module. The CDP is learned from the pairs of HR and LR images in the first stage, while recreated from the LR images solely via a diffusion-based estimator in the second stage.
    \item We propose a CDP-guided SR network where CDP is injected via learnable affine transformations as well as interflow mechanisms to improve the representation of both high- and low-frequency details.
    \item Experiments demonstrate the superiority of CDFormer, leading to a new state-of-the-art performance. With content estimation, CDFormer achieves unprecedented SR results even for severely degraded images.
\end{itemize}


%% file: sec/2_relwork.tex
\section{Related Work}
\label{sec:related}

\subsection{Image Super-Resolution}
\textbf{Non-blind Image Super-Resolution} algorithms, pioneered by SRCNN \cite{8954253}, mostly start with an assumption in the degradation process (\eg bicubic downsampling and specific blur kernels) and can produce appealing SR results for the synthetic cases, employing either recurrent neural networks (RNNs) \cite{7780550,8099781}, adversarial learning \cite{ledig2017photo,9607421,9878729} or attention mechanisms \cite{8954252,liu2018non,9578003,niu2020single}. To meet the real-world challenges with multiple degrading effects, SRMD \cite{8954253} proposes to incorporate an LR image with a stretched blur kernel, and UDVD \cite{9156922} utilizes consecutive dynamic convolutions and multi-stage loss to refine image features.  Recently, Transformer-based networks \cite{9577359,liang2021swinir,chen2023dual} have emerged and achieved state-of-the-art results. Nevertheless, most non-blind image Super-Resolution methods fail to cope with complicated degradation cases.
\\
\textbf{Blind Image Super-Resolution} then emerged to deal with real-world scenarios where the degradation kernels are complicated. Previous methods based on Kernel Prediction (KP) utilize explicit \cite{9156619, gu2019blind, 9578133, 9157144, 9157092} or implicit \cite{bell2019blind, 9879171} techniques to predict blur kernels, and then guide non-blind SR networks through kernel stretching strategies. DCLS \cite{9879171} presents a least squares deconvolution operator to produce clean features using estimated kernels. While KP methods require the ground truth labels of the degradation kernels, Degradation Prediction (DP) methods further encourage learning degradation representation. DASR \cite{wang2021unsupervised} extracts degradation representation in an unsupervised manner. KDSR \cite{xia2022knowledge} takes full advantage of knowledge distillation to distinguish different degradations. 

Although the above DP methods can handle real-world degradations to some extent, we do emphasize that only degradation representation has been utilized in previous works, which may render the network agnostic to the texture information. In contrast, our proposed CDFormer explicitly takes both degradation and content representations into consideration and can achieve remarkable improvement. 

\subsection{Diffusion Models for Super-Resolution}
Diffusion models (DMs) \cite{ho2020denoising, kingma2021variational, dhariwal2021diffusion} have been proving to be a powerful generative engine. Compared to other generative models like Generative Adversarial Networks (GANs), DMs define a parameterized Markov chain to optimize the lower variational bound on the likelihood function, allowing it to fit all data distributions. DMs for the super-resolution task have been investigated in several papers. SR3 \cite{9887996} and SRDiff \cite{li2022srdiff} adopt DDPM \cite{ho2020denoising} for SR to reconstruct the LR image via iterative denoising, StableSR \cite{wang2023exploiting} further explore the Latent Diffusion Model (LDM) \cite{rombach2022high} to enhance efficiency. However, as discussed above, the use of DMs as SR net can induce error propagation and consequently incorrect textures in SR results. Moreover, DMs are known for being quite expensive in terms of time and computation, which is crucial for real-time applications. 

The solution of generating particular information by diffusion models has provided a new perspective to enhance accuracy and stability. DiffIR \cite{10377629} that trains the DM to estimate prior representation achieves state-of-the-art performance in image restoration. More than DiffIR, we encourage the diffusion process as an estimator to recover both degradation and content representations from the LR images, therefore improving reconstruction in textural details.

%% file: sec/3_method.tex
\section{Methodology}
\begin{figure*}[t]
    \vspace{-1em}
	\centering
	\includegraphics[width=1.0\linewidth]{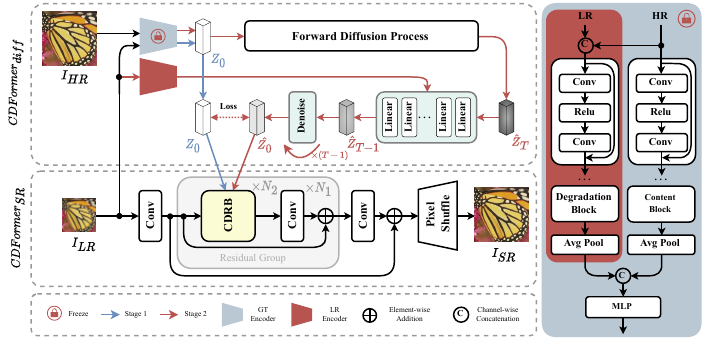}
	\caption{Overall architecture of our proposed CDFormer. In the first stage (blue line), we train the GT encoder to learn Content Degradation Prior (CDP) from both HR and LR images to guide the SR network $CDFormer_{SR}$. In the second stage (red line), only LR images are input into LR encoder to produce conditional vectors, which helps the diffusion model to recreate CDP.}
    \vspace{-1em}
	\label{fig:main}
\end{figure*}
\subsection{Motivation}
Previous BSR methods based on KP have been observed to be ineffective when dealing with complex degradations. DP approaches instead estimate degradation representation, while still in a content-independent manner. To address these limitations, we propose to predict both degradation and content representations, thus reconstructing images with more harmonious textures. 

The content detail is supposed to be rich in high-resolution images. However, SR task given only low-resolution images naturally lacks such information. In this case, we adopt the diffusion model to retrace content and degradation representations from LR images. The diffusion module in CDFormer is to estimate high-level information rather than reconstruct low-level images, thus we can overcome the limitations of low efficiency and excessive diversity that existed in previous diffusion-based SR approaches. 

Moreover, recent research has shown the robust modeling capabilities of Transformers compared with pure CNN architectures, however, still lacks an inductive bias for modeling low-frequency information \cite{chen2023dual}. We are motivated to redesign an adaptive SR network that takes the full advantage of the estimated CDP to model both high- and low-frequency information for accurate reconstruction.
\subsection{Our Method}
As illustrated in \cref{fig:main}, our proposed method is composed of a Content Degradation Prior generation module $CDFormer_{diff}$ and a Content-aware Degradation-driven Transformer SR module $CDFormer_{SR}$. We apply a two-stage training strategy to steer the diffusion model. \cref{stage1} describes the training of ground-truth encoder $E_{GT}$ that learns CDP from both LR and HR images. \cref{stage2} presents the pipeline to generate CDP from LR images by encoder $E_{LR}$ with the diffusion process. In both stages, $CDFormer_{SR}$ is trainable, which is stacked by some residual groups and a reconstruction net. Each residual group introduces several Content-aware Degradation-driven Refinement Blocks (CDRBs), where Content Degradation Injection Module (CDIM) is applied to infuse the estimated CDP. 
For simplicity, we provide a brief background of the diffusion model in the supplementary material and the following will focus on our approach.
\subsubsection{STAGE 1: Learn CDP from $I_{LR}$ and $I_{HR}$}
\label{stage1}
In this stage, $E_{GT}$  will be trained to construct the real data distribution under the supervision of both HR and LR images, and $CDFormer_{SR}$ will also be trained to ensure that estimated representations have been efficiently used.

The estimator $E_{GT}$ is designed to extract the degradation information from LR and HR images pair, as well as the content details from the HR images, which is the so-called CDP, denoted as:
\begin{equation}
	Z_{0}=E_{GT}(Concat((I_{HR})\downarrow_s, I_{LR}), I_{HR}),
	\label{eq:Encoder_GT}
\end{equation}
where the CDP $Z_{0}\in \mathbb{R}^{C_z}$, $\downarrow_s$ denotes a Pixel-Unshuffle operation with scale $s$. Specifically, $E_{GT}$ is implemented by several residual blocks and an MLP layer, as depicted in the right of \cref{fig:main}. The fusion of CDP with features in $CDFormer_{SR}$ is accomplished in the injection module CDIM through channel-wise affine transformations which are learnable during training, formulated as:
\begin{equation}
	F^{'}=Linear(Z_{0})\odot Norm(F)+Linear(Z_{0}),
	\label{eq:CDIM_ori}
\end{equation}
where $F,F^{'}\in \mathbb{R}^{H\times W\times C}$ are input and output feature maps respectively, $\odot$ is a element-wise multiplication, $Norm(\cdot)$ denotes Layer Normalization.

The goal of SR network $CDFormer_{SR}$ is to reconstruct high-resolution images with the guidance of CDP.  
To enhance the representation ability, CDRB combines spatial attention and channel attention mechanisms, and further incorporates CNN and Transformer features via interflow mechanism, enabling the CDRB module to adaptively refine both high and low-frequency information.

\begin{figure*}[t]
	\centering
	\includegraphics[width=0.98\linewidth]{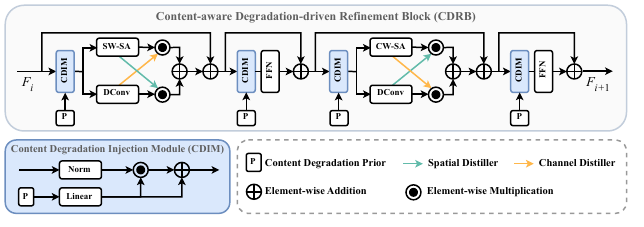}
    \vspace{-10pt}
	\caption{Details of Content-aware Degradation-driven Refinement Block (CDRB).}
    \vspace{-1em}
	\label{fig:details}
\end{figure*}

As depicted in \cref{fig:details}, $i$-th CDRB involves four CDIMs to inject CDP into the feature maps $F^{j}_{i} \in\mathbb{R}^{H\times W\times C}$ by:
\begin{equation}
	\hat{F}^{j}_{i}=CDIM_{i,j}(F^{j}_{i}, Z_{0}),i=1,...,N_{2}, j=1,2,3,4
	\label{eq:CDIM}
\end{equation}
where $Z_{0}$ is the CDP predicted by $E_{GT}$, $CDIM_{i,j}$ is the $j$-th CDIM in $i$-th CDRB. 

To ensure effective learning of representations, we apply two kinds of self-attention and deep convolution operations. To be specific, we first utilize Spatial Window Self-Attention (SW-SA) that calculates attention scores within non-overlapping windows. Given the input features $\hat{F}^{1}_{i}=CDIM_{i,1}(F^1_i,Z_0)$, we first obtain query, key, and value elements through linear projection: 
\begin{equation}
	Q=\hat{F}^{1}_{i} W_{Q},K=\hat{F}^{1}_{i} W_{K},V=\hat{F}^{1}_{i} W_{V},
\end{equation}
where $W_Q,W_K,W_V\in \mathbb R^{C\times C}$ are learnable parameter matrices, with no bias appended. Window partition is then applied to divide $Q,K,V$ into $\frac{HW}{N_{w}}$ non-overlapping windows, each window with $N_w$ length. These flattened and reshaped elements are denoted as $Q_s,K_s$, and $V_{s}$. Finally, the output features are obtained as follows:
\begin{equation}
    F^{S}_i=Softmax({Q_{s}K_{s}^T}/{\sqrt{d_k}}) V_{s},
	\label{eq:SWSA}
\end{equation}
where $d_k$ is the variance of the attention score.

Moreover, we introduce an interflow mechanism that utilizes two kinds of distiller between two branches to adaptively modulate the CNN and Transformer features. Specifically, for feature maps in $\mathbb{R}^{H\times W\times C}$, Channel Distiller $\Tilde{C}$ transforms to $\mathbb{R}^{1\times 1\times C}$, while Spatial Distiller $\Tilde{S}$ changes to $\mathbb{R}^{H\times W\times 1}$. We formulate the above process as:
\begin{equation}
\begin{aligned}
	&F^{S}_i=SW\mbox{-}SA(\hat{F}^{1}_{i}), F^{W}_i=DConv(\hat{F}^{1}_{i}), \\
	&F^{2}_{i}=F^{S}_i \odot \Tilde{S}(F^{W}_i) + F^{W}_i \odot \Tilde{C}(F^{S}_i)+F^1_{i},\\
    &F^{3}_{i}=FFN(\hat{F}^{2}_{i})+F^{2}_{i}
\label{eq:FAS}
\end{aligned}
\end{equation}
where $DConv(\cdot)$ denotes a depth-wise convolution layer, $\hat{F}^{2}_{i}=CDIM_{i,2}(F_i^2,Z_0)$, $FFN(\cdot)$ is a feed-forward network with GELU activation.

Different from SW-SA which learns pixel-wise relations, Channel-wise Self-Attention (CW-SA) focuses on understanding relationships between channels. Given feature maps after injection operation as $\hat{F}^{3}_{i}=CDIM_{i,3}(F^3_i,Z_0)$, query, key, and value elements $Q_c, K_c, V_c$ are projected from $\hat{F}^{3}_{i}$. Channel-wise relationship is computed by:
\begin{equation}
F^{C}_i=Softmax({Q_{c}^TK_{c}}/{\alpha})V_{c},
	\label{eq:CASA}
\end{equation}
where $\alpha$ is a learnable temperature parameter.

Again, feature aggregation is achieved by introducing the interflow between channel-wise self-attention and deep convolution operations. This technique allows the network to capture both global and local dependencies. This process is formulated as follows:
\begin{equation}
\begin{aligned}
	&F^{C}_i= CW\mbox{-}SA(\hat{F}^{3}_{i}),\dot{F}^{W}_i=DConv(\hat{F}^{3}_{i}), \\
	&F^{4}_{i}= F^{C}_i \odot \Tilde{C}(\dot{F}^{W}_i)+\dot{F}^{W}_i \odot \Tilde{S}(F^{C}_i)+F^3_{i},\\
    &F_{i+1}=FFN(\hat{F}^{4}_{i})+F^{4}_{i}
	 \label{eq:FAC}
\end{aligned}
\end{equation}
where $\Tilde{C}$ and $\Tilde{S}$ denote the channel distiller and spatial distiller, respectively. $\hat{F}^{4}_{i}=CDIM_{i,4}(F^4_i,Z_0)$.

Finally, we jointly optimize $E_{GT}$ and $CDFormer_{SR}$ by reconstruction loss between HR images and reconstructed SR images as the training objective for the first stage:
\begin{equation}
	\mathcal {L}_{rec}=\|I_{HR}-I_{SR}\|_{1},
	\label{eq:S1loos}
\end{equation}

\subsubsection{STAGE 2: Generate CDP from $I_{LR}$}
\label{stage2}
Theoretically, $I_{HR}$ is full of content details. However, for the SR task, such high-resolution images are unknown during inference. In the second stage, we therefore propose to treat the diffusion model as an estimator and generate CDP from LR images, exploiting the capability of diffusion models to approximate the real data distribution.

Specifically, we reuse the pre-trained encoder $E_{GT}$ in the first stage to produce the initial CDP $Z_{0}\in \mathbb{R}^{C_z}$, which is supposed to be the ground truth representation of content and degradation. Following the normal routine to train a diffusion model, we add Gaussian noise to $Z_0$ in the forward diffusion process to produce a noisy representation:
\begin{equation}
	Z_T = \sqrt{\bar \alpha_T} Z_0 + \sqrt{1-\bar\alpha_T} \epsilon
\end{equation}
where $T$ is the total number of time steps, $\bar\alpha_T=\prod_{i=1}^T{\alpha_i}$ is predefined schedule, and added noise $\epsilon \sim \mathcal{N}(0,1)$.

During the reverse process, $E_{LR}$ that is a replicate of $E_{GT}$ with the degradation branch only, will be trained to produce a one-dimensional condition vector $c\in \mathbb{R}^{C_z}$ from the LR images. This conditional vector will then be fused into each reverse step, guiding the diffusion model to generate a great representation of CDP based on the LR images. Notice that each training step $CDFormer_{diff}$ performs the whole sampling process with $T$ iterations, which is different from the traditional DMs that minimize $||\epsilon-\epsilon_\theta(x_t,t,c)||$ for a single diffusion step. Instead, we compute $\hat{Z}_0$ in every training step as follows:
\begin{equation}
\begin{aligned}
    \hat{Z}_{t-1}=\dfrac{1}{\sqrt{\alpha_{t}}}(\hat{Z}_{t}-\dfrac{1-\alpha_{t}}{\sqrt{1-\bar\alpha_{t}}}\epsilon_\theta(\hat{Z}_t,t,c))+\sigma_t\epsilon,\\
    t=T,...,1
\label{eq:reverse}
\end{aligned}
\end{equation}
where the one-dimensional condition vector $c=E_{LR}(I_{LR})$ is concatenated with $\hat{Z}_t$, variance $\sigma_t=\sqrt{1-\alpha_{t}}$.

The generated representation $\hat{Z}_0$ is expected to recreate the lost information in LR images, thus we compute the $L_1$ distance between the CDP from $E_{GT}$ and from diffusion as:
\begin{equation}
	\mathcal {L}_{diff}=\|Z_{0}-\hat{Z_{0}}\|_{1}	\label{eq:loss}
\end{equation}

To ensure efficient learning, $CDFormer_{diff}$ and $CDFormer_{SR}$ will be trained jointly, by minimizing the loss function $\mathcal {L}= \mathcal L_{diff}+ \alpha_{rec} \mathcal L_{rec}$. Notice that in the second stage, we inject CDP $\hat{Z}_0$ predicted by diffusion model into $CDFormer_{SR}$, instead of $Z_0$ in the first stage. The algorithm is detailed in the supplementary material.

During training, $\hat{Z}_T$ is actually given from $I_{HR}$, while for inference we exclusively perform the reverse diffusion process from a Gaussian noise, i.e., $\hat{Z}_T \sim \mathcal{N}(0,1)$. The conditional vector $c$ obtained from LR images will also participate in. Utilizing the denoising ability of the diffusion model, CDFormer is able to generate CDP from LR images, with abundant content and degradation representations to guide $CDFormer_{SR}$ to reconstruct both high- and low-frequency information. Moreover, $CDFormer_{diff}$ achieves plausible results with fewer sampling iterations ($T=4$ in our experiments) and parameters ($\sim 3$M).

%% file: sec/4_experiments.tex
\section{Experiments}
\begin{table*}[t]
	\caption{Comparison with diffusion-based SR models test on $192\times192$ resolution for $\times 4$ scale. KP method DCLS for reference.}
    \vspace{-6pt}
	\centering
	\label{table:dms}
	\resizebox{\textwidth}{!}{
		\begin{tabular}{|c|ccc|cccc|cccc|}
			\hline
			\multirow{2}{*}{Method} & 	\multirow{2}{*}{Params(M)} & 	\multirow{2}{*}{FLOPs(T)} &	
			\multirow{2}{*}{Time(s)} &	  \multicolumn{4}{c}{Set14 / kernel width=0}&  \multicolumn{4}{|c|}{Urban100 / kernel width=0}  \\
			&&&& PSNR $\uparrow$ & SSIM $\uparrow$  & FID $\downarrow$ & LPIPS $\downarrow$ & PSNR $\uparrow$ & SSIM $\uparrow$  & FID $\downarrow$ & LPIPS $\downarrow$   \\
			\hline
			SR3\cite{9887996}   & 187.65 & 6601 & 532 &
			25.41 & 0.7462 & 83.63 & \textbf{\textcolor{red}{0.246}} & - & - & - & -  \\
			StableSR \cite{wang2023exploiting}  & 1409.11 & - & 71 &
			19.66 & 0.5406 & 97.84 & 0.282 & 20.56 & 0.6753 & 54.21 & \textbf{\textcolor{red}{0.195}}  \\
   		DCLS\cite{9879171}  & 13.63 & 0.279 & 0.04 &
			28.61 & 0.7816 & \textcolor{blue}{\underline{78.21}} & 0.293 & 26.50 & 0.7973 & 43.25 & 0.220  \\
			CDFormer-S  & 11.09 & 0.355 & 0.14 &
			\textcolor{blue}{\underline{28.80}} & \textcolor{blue}{\underline{0.7876}} & 78.69 & 0.287 & \textcolor{blue}{\underline{26.56}} & \textcolor{blue}{\underline{0.8023}} & \textcolor{blue}{\underline{26.40}} & 0.214 \\ 
			CDFormer  & 24.46 & 0.725 & 0.39 &
			\textbf{\textcolor{red}{29.00}} & \textbf{\textcolor{red}{0.7918}} & \textbf{\textcolor{red}{75.43}} & \textcolor{blue}{\underline{0.280}} & \textbf{\textcolor{red}{27.21}} & \textbf{\textcolor{red}{0.8189}} & \textbf{\textcolor{red}{23.94}} & \textcolor{blue}{\underline{0.197}}  \\
			\hline
	\end{tabular}}
    \vspace{-6pt}
\end{table*}
\begin{table*}[t]
	\caption{Quantitive evaluation by PSNR ($\uparrow$) on noise-free degradations with isotropic Gaussian kernels. Best in \textbf{\textcolor{red}{red}} and second in \textcolor{blue}{\underline{blue}}.}
    \vspace{-6pt}
	\centering
	\label{table:iso}
	\resizebox{\textwidth}{!}{
		\begin{tabular}{|c|c|cccc|cccc|cccc|cccc|}
			\hline
			Method&
			Scale&
			\multicolumn{4}{c|}{Set5}& \multicolumn{4}{c|}{Set14}& \multicolumn{4}{c|}{B100}& \multicolumn{4}{c|}{Urban100} \\
            \hline
			\multicolumn{2}{|c|}{Kernel Width} &  0   &  0.6  &  1.2  &  1.8  &   0   &  0.6  &  1.2  &  1.8  &   0   &  0.6  &  1.2  &  1.8  &   0   &  0.6  &  1.2  &  1.8   \\
			\hline
			Bicubic & \multirow{5}{*}{$\times$2}      & 33.66 & 32.30 & 29.28 & 27.07 & 30.24 & 29.21 & 27.13 & 25.47 & 29.56 & 28.76 & 26.93 & 25.51 & 26.88 & 26.13 & 24.46 & 23.06  \\
			RCAN \cite{zhang2018image} &              & \textbf{\textcolor{red}{38.27}} & 35.91 & 31.20 & 28.50 & 
			\textbf{\textcolor{red}{34.12}} & 32.31 & 28.48 & 26.33 & \textbf{\textcolor{red}{32.41}} & 31.16 & 28.04 & 26.26 & 
			\textbf{\textcolor{red}{33.34}} & 29.80 & 25.38 & 23.44  \\
			DASR \cite{wang2021unsupervised} &              & 
			37.87 & 37.47 & 37.19 & 35.43 & 
			33.34 &32.96 & 32.78 &31.60 & 
			32.03 & 31.78 & 31.71 & 30.54 &
			31.49 & 30.71 & 30.36 & 28.95  \\
			DCLS \cite{9879171} &              & 
			38.06 & \textcolor{blue}{\underline{38.04}} & \textcolor{blue}{\underline{37.66}} & \textcolor{blue}{\underline{36.06}} & 
			33.62 & \textcolor{blue}{\underline{33.52}} & \textcolor{blue}{\underline{33.52}} & \textcolor{blue}{\underline{32.27}} & 
			32.23 & \textcolor{blue}{\underline{32.25}} & \textcolor{blue}{3\underline{2.12}} & \textcolor{blue}{\underline{30.93}} & 
			32.36 & \textcolor{blue}{\underline{32.17} }& \textcolor{blue}{\underline{31.81}} & \textcolor{blue}{\underline{30.23}}  \\ 
			\multirow{1}{*}{CDFormer (Ours)}    &          &
			\textcolor{blue}{\underline{38.25}} & \textbf{\textcolor{red}{38.25}} & \textbf{\textcolor{red}{37.88}} & \textbf{\textcolor{red}{36.32}} &
			\textcolor{blue}{\underline{34.10}} & \textbf{\textcolor{red}{34.01}} & \textbf{\textcolor{red}{33.88}} & \textbf{\textcolor{red}{32.57}}&
			\textcolor{blue}{\underline{32.40}} & \textbf{\textcolor{red}{32.39}} & \textbf{\textcolor{red}{32.25}} & \textbf{\textcolor{red}{31.03}} &
			\textcolor{blue}{\underline{33.11}} & \textbf{\textcolor{red}{32.62}} & \textbf{\textcolor{red}{32.08}} & \textbf{\textcolor{red}{30.47}}     \\
			\hline
			\multicolumn{2}{|c|}{Kernel Width} &   0   &  0.8  &  1.6  &  2.4  &   0   &  0.8  &  1.6  &  2.4  &   0   &  0.8  &  1.6  &  2.4  &   0   &  0.8  &  1.6  &  2.4   \cr\hline
			Bicubic & \multirow{5}{*}{$\times$3}      & 30.39 & 29.42 & 27.24 & 25.37 & 27.55 & 26.84 & 25.42 & 24.09 & 27.21 & 26.72 & 25.52 & 24.41 & 24.46 & 24.02 & 22.95 & 21.89  \\
			RCAN \cite{zhang2018image} &              & \textcolor{blue}{\underline{34.74}} & 32.90 & 29.12 & 26.75 & 
			\textcolor{blue}{\underline{30.65}} & 29.49 & 26.75 & 24.99 & \textcolor{blue}{2\underline{9.32}} & 28.56 & 26.55 & 25.18 & 
			\textcolor{blue}{\underline{29.09}} & 26.89 & 26.89 & 22.30  \\
			DASR \cite{wang2021unsupervised} &              &
			34.06 &34.08 & 33.57 & 32.15& 
			30.13 &29.99 &28.66 & 28.42 & 
			28.96 &28.90 &28.62 & 28.13 &
			27.65 & 27.36 & 26.86 & 25.95 \\
			DCLS \cite{9879171} &              & 
			34.62 & \textcolor{blue}{\underline{34.68}} & \textcolor{blue}{\underline{34.53}} & \textcolor{blue}{\underline{33.55}} & 
			30.33 & \textcolor{blue}{\underline{30.39}} & \textcolor{blue}{\underline{30.42}} & \textcolor{blue}{\underline{29.76}} & 
			29.16 & \textcolor{blue}{\underline{29.21}} & \textcolor{blue}{\underline{29.20}} & \textcolor{blue}{\underline{28.68}} & 
			28.53 & \textcolor{blue}{\underline{28.50}} & \textcolor{blue}{\underline{28.29}} & \textcolor{blue}{\underline{27.47}}  \\
			\multirow{1}{*}{CDFormer (Ours)}    &             &
			\textbf{\textcolor{red}{34.79}} & \textbf{\textcolor{red}{34.85}} & \textbf{\textcolor{red}{34.61}} & \textbf{\textcolor{red}{33.73}} &
			\textbf{\textcolor{red}{30.73}} & \textbf{\textcolor{red}{30.70}} & \textbf{\textcolor{red}{30.60}} & \textbf{\textcolor{red}{29.94}} &
			\textbf{\textcolor{red}{29.34}} & \textbf{\textcolor{red}{29.36}} & \textbf{\textcolor{red}{29.30}} & \textbf{\textcolor{red}{28.79}} &
			\textbf{\textcolor{red}{29.20}} & \textbf{\textcolor{red}{29.01}} & \textbf{\textcolor{red}{28.68}} & \textbf{\textcolor{red}{27.86}}     \\		
			
            \hline
			\multicolumn{2}{|c|}{Kernel Width} &   0   &  1.2  &  2.4  &  3.6  &   0   &  1.2  &  2.4  &  3.6  &   0   &  1.2  &  2.4  &  3.6 &   0   &  1.2  &  2.4  &  3.6    \cr \hline
			Bicubic & \multirow{7}{*}{$\times$4}      & 28.42 & 27.30 & 25.12 & 23.40 & 26.00 & 25.24 & 23.83 & 22.57 & 25.96 & 25.42 & 24.20 & 23.15 & 23.14 & 22.68 & 21.62 & 20.65  \\
			RCAN \cite{zhang2018image} &              & \textcolor{blue}{\underline{32.63}} & 30.26 & 26.72 & 24.66 & 
			28.87 & 27.48 & 24.93 & 23.41 & 27.72 & 26.89 & 25.09 & 23.93 & 26.61 & 24.71 & 22.25 & 20.99  \\
			DASR \cite{wang2021unsupervised} &              & 31.99 & 31.92 & 31.75 & 30.59 & 
			28.50 & 28.45 & 28.28 & 27.45 & 
			27.51 & 27.52 & 27.43 & 26.83 &
			25.82 &25.69 & 25.44 & 24.66  \\
			KDSR \cite{xia2022knowledge} &              &
			32.42 &  32.34 &  32.13 &  31.02 &
			28.67 & 28.66 &  28.55 &  27.80 &
			27.64 &  27.67 &  27.60 &  26.98 &
			26.36 &  26.29 &  26.06 &  25.21 \\
			
			KDSR-L \cite{xia2022knowledge} &              &
			32.46 &  \textcolor{blue}{\underline{32.46}} &  \textcolor{blue}{\underline{32.22}} &  \textcolor{blue}{\underline{31.17}} &
			\textcolor{blue}{\underline{28.99}} & \textbf{\textcolor{red}{29.02}} &  \textbf{\textcolor{red}{28.79}} &  \textbf{\textcolor{red}{27.94}} &
			\textcolor{blue}{\underline{27.78}} &  \textcolor{blue}{\underline{27.83}} &  \textcolor{blue}{\underline{27.73}} &  \textcolor{blue}{\underline{27.07}} &
			\textcolor{blue}{\underline{26.62}} &  \textcolor{blue}{\underline{26.58}} &  \textcolor{blue}{\underline{26.25}} &  \textcolor{blue}{\underline{25.38}} \\
			DCLS \cite{9879171}  &              &
			32.36 & 32.35 & 32.19 & 31.14 & 
			28.61 & 28.66 & 28.57 & 27.78& 
			27.69 & 27.73 & 27.65 & 27.02 & 
			26.50 & 26.50 & 26.24 & 25.34  \\

			\multirow{1}{*}{CDFormer (Ours)}    &             &
			\textbf{\textcolor{red}{32.69}} & \textbf{\textcolor{red}{32.65}} & \textbf{\textcolor{red}{32.24}} & \textbf{\textcolor{red}{31.33}} &
			\textbf{\textcolor{red}{29.00}} & \textcolor{blue}{\underline{28.99}} & \textcolor{blue}{\underline{28.75}} & \textcolor{blue}{\underline{27.93}} &
			\textbf{\textcolor{red}{27.86}} & \textbf{\textcolor{red}{27.86}} & \textbf{\textcolor{red}{27.76}} & \textbf{\textcolor{red}{27.12}} &
			\textbf{\textcolor{red}{27.21}} & \textbf{\textcolor{red}{27.06}} & \textbf{\textcolor{red}{26.66}} & \textbf{\textcolor{red}{25.72}}     \\
			\hline
		\end{tabular}
	}
 \vspace{-6pt}
\end{table*}
\begin{figure*}[!htbp]
	\centering
	\begin{subfigure}[b]{0.22\textwidth}
	\centering
	\includegraphics[width=1.0\textwidth]{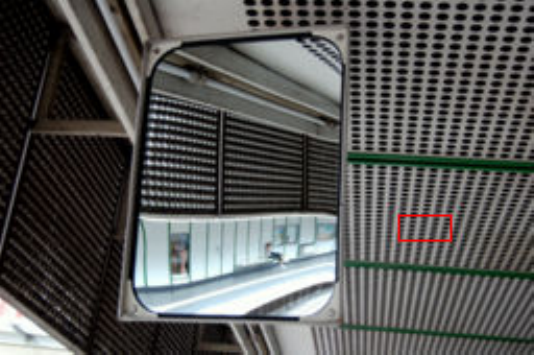}
	\subcaption*{LR Img 4 in Urban100}
\end{subfigure}
\begin{subfigure}[b]{0.27\textwidth}
	\centering
	\begin{subfigure}[b]{0.3\textwidth}
		\centering
		\includegraphics[width=1.0\textwidth]{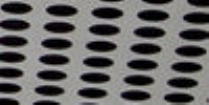}
		\subcaption*{\makebox[1em][c]{GT} \\ \makebox[1em][c]{PSNR/SSIM}}
	\end{subfigure}
	\begin{subfigure}[b]{0.3\textwidth}
		\centering
		\includegraphics[width=1.0\textwidth]{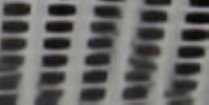}
		\subcaption*{\makebox[1em][c]{DASR} \\ \makebox[1em][c]{23.17/0.762}}
	\end{subfigure}
	\begin{subfigure}[b]{0.3\textwidth}
		\centering
		\includegraphics[width=1.0\textwidth]{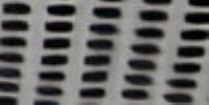}
		\subcaption*{\makebox[1em][c]{KDSR} \\ \makebox[1em][c]{24.04/0.842}}
	\end{subfigure}
	\begin{subfigure}[b]{0.3\textwidth}
		\centering
		\includegraphics[width=1.0\textwidth]{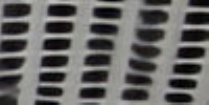}
		\subcaption*{\makebox[1em][c]{DCLS} \\ \makebox[1em][c]{24.00/0.855}}
	\end{subfigure}
	\begin{subfigure}[b]{0.3\textwidth}
		\centering
		\includegraphics[width=1.0\textwidth]{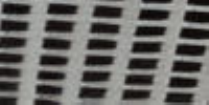}
		\subcaption*{\makebox[1em][c]{StableSR} \\ \makebox[1em][c]{18.85/0.64}}
	\end{subfigure}
	\begin{subfigure}[b]{0.3\textwidth}
		\centering
		\includegraphics[width=1.0\textwidth]{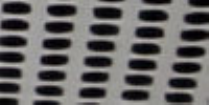}
		\subcaption*{\makebox[1em][c]{Ours} \\ \makebox[1em][c]{\textbf{\red{25.53/0.883}}}}
	\end{subfigure}
\end{subfigure}
	\begin{subfigure}[b]{0.22\textwidth}
	\centering
	\includegraphics[width=1.0\textwidth]{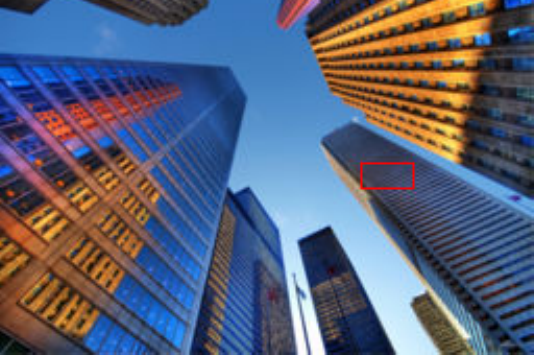}
	\subcaption*{LR Img 12 in Urban100}
\end{subfigure}
\begin{subfigure}[b]{0.27\textwidth}
	\centering
	\begin{subfigure}[b]{0.3\textwidth}
		\centering
		\includegraphics[width=1.0\textwidth]{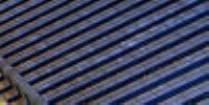}
		\subcaption*{\makebox[1em][c]{GT} \\ \makebox[1em][c]{PSNR/SSIM}}
	\end{subfigure}
	\begin{subfigure}[b]{0.3\textwidth}
		\centering
		\includegraphics[width=1.0\textwidth]{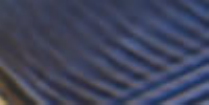}
		\subcaption*{\makebox[1em][c]{DASR} \\ \makebox[1em][c]{23.68/0.713}}
	\end{subfigure}
	\begin{subfigure}[b]{0.3\textwidth}
		\centering
		\includegraphics[width=1.0\textwidth]{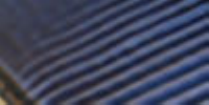}
		\subcaption*{\makebox[1em][c]{KDSR} \\ \makebox[1em][c]{23.91/0.736}}
	\end{subfigure}
	\begin{subfigure}[b]{0.3\textwidth}
		\centering
		\includegraphics[width=1.0\textwidth]{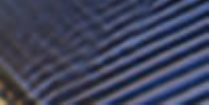}
		\subcaption*{\makebox[1em][c]{DCLS} \\ \makebox[1em][c]{24.25/0.744}}
	\end{subfigure}
	\begin{subfigure}[b]{0.3\textwidth}
		\centering
		\includegraphics[width=1.0\textwidth]{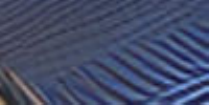}
		\subcaption*{\makebox[1em][c]{StableSR} \\ \makebox[1em][c]{19.08/0.60}}
	\end{subfigure}
	\begin{subfigure}[b]{0.3\textwidth}
		\centering
		\includegraphics[width=1.0\textwidth]{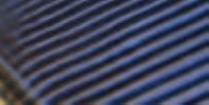}
		\subcaption*{\makebox[1em][c]{Ours} \\ \makebox[1em][c]{\textbf{\red{24.59/0.774}}}}
	\end{subfigure}
 
\end{subfigure}
	\begin{subfigure}[b]{0.22\textwidth}
	\centering
	\includegraphics[width=1.0\textwidth]{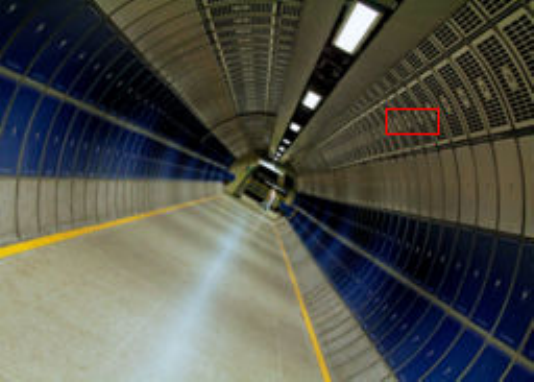}
	\subcaption*{LR Img 78 in Urban100}
\end{subfigure}
\begin{subfigure}[b]{0.27\textwidth}
	\centering
	\begin{subfigure}[b]{0.3\textwidth}
		\centering
		\includegraphics[width=1.0\textwidth]{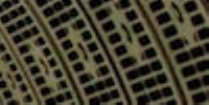}
		\subcaption*{\makebox[1em][c]{GT} \\ \makebox[1em][c]{PSNR/SSIM}}
	\end{subfigure}
	\begin{subfigure}[b]{0.3\textwidth}
		\centering
		\includegraphics[width=1.0\textwidth]{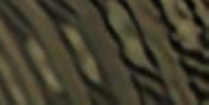}
		\subcaption*{\makebox[1em][c]{DASR} \\ \makebox[1em][c]{27.13/0.758}}
	\end{subfigure}
	\begin{subfigure}[b]{0.3\textwidth}
		\centering
		\includegraphics[width=1.0\textwidth]{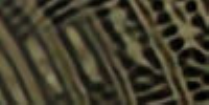}
		\subcaption*{\makebox[1em][c]{KDSR} \\ \makebox[1em][c]{27.69/0.778}}
	\end{subfigure}
	\begin{subfigure}[b]{0.3\textwidth}
		\centering
		\includegraphics[width=1.0\textwidth]{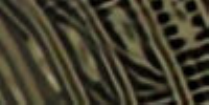}
		\subcaption*{\makebox[1em][c]{DCLS} \\ \makebox[1em][c]{28.15/0.784}}
	\end{subfigure}
	\begin{subfigure}[b]{0.3\textwidth}
		\centering
		\includegraphics[width=1.0\textwidth]{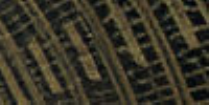}
		\subcaption*{\makebox[1em][c]{StableSR} \\ \makebox[1em][c]{23.53/0.59}}
	\end{subfigure}
	\begin{subfigure}[b]{0.3\textwidth}
		\centering
		\includegraphics[width=1.0\textwidth]{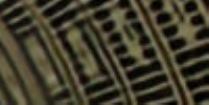}
		\subcaption*{\makebox[1em][c]{Ours} \\ \makebox[1em][c]{\textbf{\red{28.79/0.808}}}}
	\end{subfigure}
\end{subfigure}
	\begin{subfigure}[b]{0.22\textwidth}
		\centering
		\includegraphics[width=1.0\textwidth]{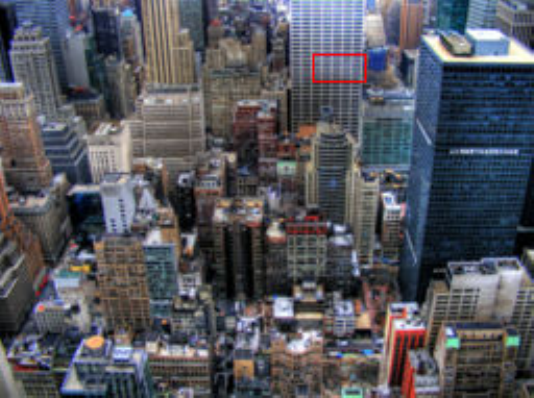}
		\subcaption*{LR Img 73 in Urban100}
	\end{subfigure}
\begin{subfigure}[b]{0.27\textwidth}
	\centering
	\begin{subfigure}[b]{0.3\textwidth}
		\centering
		\includegraphics[width=1.0\textwidth]{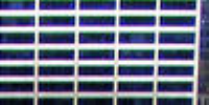}
		\subcaption*{\makebox[1em][c]{GT} \\ \makebox[1em][c]{PSNR/SSIM}}
	\end{subfigure}
	\begin{subfigure}[b]{0.3\textwidth}
		\centering
		\includegraphics[width=1.0\textwidth]{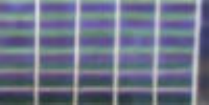}
		\subcaption*{\makebox[1em][c]{DASR} \\ \makebox[1em][c]{20.09/0.530}}
	\end{subfigure}
	\begin{subfigure}[b]{0.3\textwidth}
		\centering
		\includegraphics[width=1.0\textwidth]{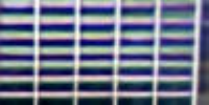}
		\subcaption*{\makebox[1em][c]{KDSR} \\ \makebox[1em][c]{20.22/0.581}}
	\end{subfigure}
	\begin{subfigure}[b]{0.3\textwidth}
		\centering
		\includegraphics[width=1.0\textwidth]{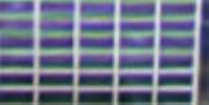}
		\subcaption*{\makebox[1em][c]{DCLS} \\ \makebox[1em][c]{20.59/0.583}}
	\end{subfigure}
	\begin{subfigure}[b]{0.3\textwidth}
		\centering
		\includegraphics[width=1.0\textwidth]{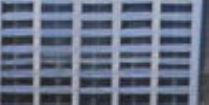}
		\subcaption*{\makebox[1em][c]{StableSR} \\ \makebox[1em][c]{16.91/0.39}}
	\end{subfigure}
	\begin{subfigure}[b]{0.3\textwidth}
		\centering
		\includegraphics[width=1.0\textwidth]{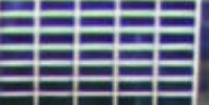}
		\subcaption*{\makebox[1em][c]{Ours} \\ \makebox[1em][c]{\textbf{\red{21.62/0.636}}}}
	\end{subfigure}
\end{subfigure}
	\caption{Visual results of Imgs in Urban100, for scale factor $4$ and kernel width $1.2$. Best marked in \textbf{\red{red}}.}
    \vspace{-1em}
	\label{fig:urban}
\end{figure*}

\begin{table*}[!t]
	\caption{Quantitive evaluation by PSNR ($\uparrow$) on Set14 for ×4 SR with anisotropic Gaussian kernels and noises. Best marked in \textbf{bold}.}
    \vspace{-6pt}
	\label{table:aniso}
	\centering
	\resizebox{0.8\textwidth}{!}{
		\begin{tabular}{|c|c|c c c c c c c c c|}
			\hline
			\multirow{3}{*}{method}                   & \multirow{3}{*}{noise} & \multicolumn{9}{c|}{Blur Kernel}\\
			&&\includegraphics[height=20pt,width=20pt]{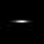}
			&\includegraphics[height=20pt,width=20pt]{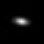}
			&\includegraphics[height=20pt,width=20pt]{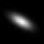}
			&\includegraphics[height=20pt,width=20pt]{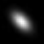}
			&\includegraphics[height=20pt,width=20pt]{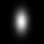}
			&\includegraphics[height=20pt,width=20pt]{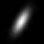}
			&\includegraphics[height=20pt,width=20pt]{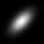} 
			&\includegraphics[height=20pt,width=20pt]{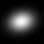} 
			&\includegraphics[height=20pt,width=20pt]{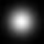}        \\
			\hline
			\multirow{3}{*}{DnCNN\cite{7839189}+RCAN\cite{zhang2018image}}               & 0                      & 26.44 & 26.22 & 24.48 & 24.23 & 24.29 & 24.19 & 23.9  & 23.42 & 23.01  \\
			& 5                      & 26.10 & 25.90 & 24.29 & 24.07 & 24.14 & 24.02 & 23.74 & 23.31 & 22.92  \\
			& 10                     & 25.65 & 25.47 & 24.05 & 23.84 & 23.92 & 23.8  & 23.54 & 23.14 & 22.77  \\\hline
			\multirow{3}{*}{DnCNN\cite{7839189}+DCLS\cite{9879171} }                      & 0                      & 27.56 & 27.49 & 26.32 & 25.99 & 25.88 & 26.03 & 25.70 & 24.65 & 23.95  \\
			& 5                      & 26.20 & 26.02 & 24.44 & 24.21 & 24.28 & 24.14 & 23.88 & 23.40 &  22.98  \\
			& 10                     & 25.47 & 25.33 & 24.06 & 23.87 & 23.91 & 23.79 & 23.58 & 23.16 & 22.78  \\\hline
			\multirow{3}{*}{DASR\cite{wang2021unsupervised}}                     & 0                      & 27.99 & 27.97 &  27.53 & 27.45 & 27.43 &  27.22 & 27.19 & 26.83 &  26.21  \\
			& 5                      & 27.25 & 27.18 &  26.37 & 26.16 &  26.09 &  25.96 & 25.85 &  25.52 &  25.04  \\
			& 10                     &  26.57 &  26.51 &  25.64 &  25.47 &  25.43 &  25.31 &  25.16 &  24.80 &  24.43  \\\hline
			\multirow{3}{*}{KDSR\cite{xia2022knowledge}}                   & 0                      & 28.26 & 28.38 & 27.98 & 27.98 & 27.94 & 27.75 & 27.69  & 27.35 & 26.52  \\
			& 5                      & 27.56 & 27.55 & 26.67 & 26.49 & 26.44 & 26.35 & 26.19 & 25.78 & 25.25  \\
			& 10                     & 26.85 & 26.79 & 25.86 & 25.70 & 25.68 & 25.59 & 25.42 & 25.06 & 24.64  \\\hline
   
			\multirow{3}{*}{CDFormer(Ours)}                     & 0                      &  
			\textbf{28.63} & \textbf{28.68} & \textbf{28.29} & \textbf{28.31} & \textbf{28.21} & \textbf{28.07} & \textbf{28.04} & \textbf{27.57} & \textbf{26.97}  \\
			& 5                      & 
			\textbf{27.76} & \textbf{27.72} & \textbf{26.79} & \textbf{26.61} & \textbf{26.59} & \textbf{26.50} & \textbf{26.35} & \textbf{25.97} & \textbf{25.51} \\
			& 10                     & 
			\textbf{27.00} & \textbf{26.92} & \textbf{25.98} & \textbf{25.80} & \textbf{25.80} & \textbf{25.70} & \textbf{25.55} & \textbf{25.22} & \textbf{24.83}  \\
			\hline	
	\end{tabular}}
	\vspace{-6pt}
\end{table*}

\label{sec:exp}
\subsection{Experiment Settings}
\textbf{Implementation Details.} CDFormer in baseline is stacked by $6$ residual groups, each containing $3$ CDRBs with window size $8\times 32$ and $180$ channels, and the channel for CDP is $C_z=256$. For a fair comparison, we develop a lightweight variant of CDFormer, denoted as CDFormer-S, where CDRB window size is reduced to $4\times 16$ and channel is reduced to $96$. We use Adam \cite{kingma2014adam} optimizer with $\beta_{1}=0.9,\beta_{2}=0.99$ to update parameters, and $\alpha_{rec}=0.01$ for the second stage training. Each stage is trained with $300$ epochs and batch size $4$. We set the initial learning rate as $1 \times 10^{-4}$, and half it every $125$ epochs. 
\\
\textbf{Datasets and Metrics.} Our models are trained on two widely used datasets: DIV2K \cite{8014883} and Flickr2K \cite{lim2017enhanced}. For testing, we adopt four benchmark datasets: Set5 \cite{bevilacqua2012low}, Set14 \cite{zeyde2012single}, B100 \cite{937655}, Urban100 \cite{7299156}. All methods will be evaluated by PSNR and SSIM \cite{1284395} unless otherwise specified. Note that, some comparison results in the following are blank, since some prior models are extremely large and they cannot run on our GPU NVIDIA RTX 4090 in some scenarios that require higher memory. 

\begin{figure*}[htbp]
	\centering
	\begin{subfigure}[b]{0.26\textwidth}
		\centering
		\includegraphics[width=1.0\textwidth]{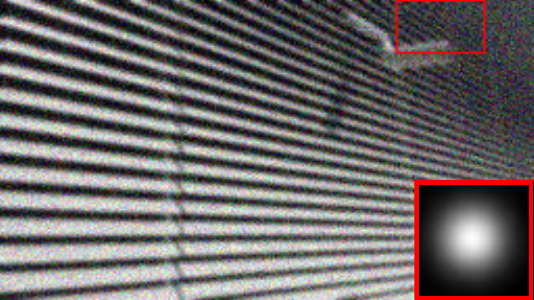}
		\subcaption*{LR in Urban100 with noise $20$.}
	\end{subfigure}
	\begin{subfigure}[b]{0.25\textwidth}
		\centering
		\begin{subfigure}[b]{0.4\textwidth}
			\centering
			\includegraphics[width=1.0\textwidth]{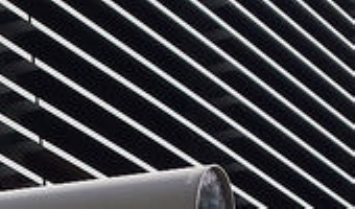}
			\subcaption*{HR}
		\end{subfigure}
		\begin{subfigure}[b]{0.4\textwidth}
			\centering
			\includegraphics[width=1.0\textwidth]{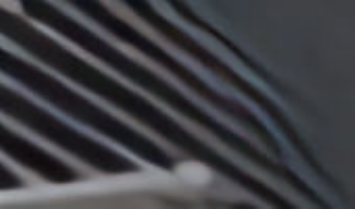}
			\subcaption*{DASR}
		\end{subfigure}
		\begin{subfigure}[b]{0.4\textwidth}
			\centering
			\includegraphics[width=1.0\textwidth]{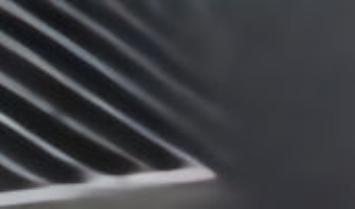}
			\subcaption*{KDSR}
		\end{subfigure}
		\begin{subfigure}[b]{0.4\textwidth}
			\centering
			\includegraphics[width=1.0\textwidth]{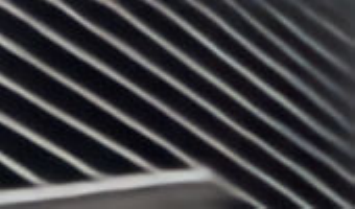}
			\subcaption*{CDFormer}
		\end{subfigure}
	\end{subfigure}
	\begin{subfigure}[b]{0.22\textwidth}
		\centering
		\includegraphics[width=1.0\textwidth]{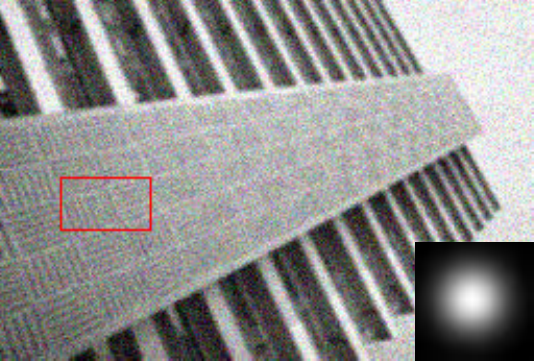}
		\subcaption*{LR in Urban100 with noise $20$.}
	\end{subfigure}
	\begin{subfigure}[b]{0.25\textwidth}
		\centering
		\begin{subfigure}[b]{0.4\textwidth}
			\centering
			\includegraphics[width=1.0\textwidth]{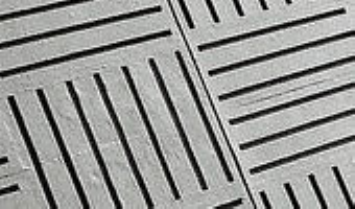}
			\subcaption*{HR}
		\end{subfigure}
		\begin{subfigure}[b]{0.4\textwidth}
			\centering
			\includegraphics[width=1.0\textwidth]{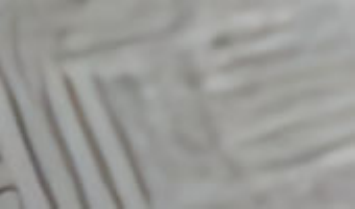}
			\subcaption*{DASR}
		\end{subfigure}
		\begin{subfigure}[b]{0.4\textwidth}
			\centering
			\includegraphics[width=1.0\textwidth]{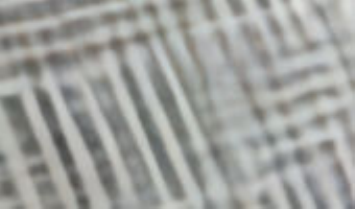}
			\subcaption*{KDSR}
		\end{subfigure}
		\begin{subfigure}[b]{0.4\textwidth}
			\centering
			\includegraphics[width=1.0\textwidth]{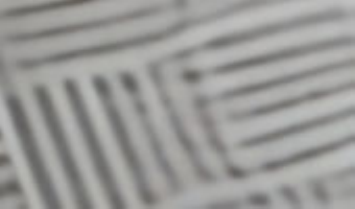}
			\subcaption*{CDFormer}
		\end{subfigure}
	\end{subfigure}
	
	\caption{Visualization of SR results via different DP methods on anisotropic Gaussian kernels and noises.}
    \vspace{-10pt}
	\label{fig:aniso}
\end{figure*}

\subsection{Comparison with State-of-the-art Methods}
\textbf{Isotropic Gaussian Kernels Noise-free Degradation}. 
We first conduct a quantitative comparison with two diffusion-based SR methods: SR3 \cite{9887996} that employs DDPM \cite{ho2020denoising} and StableSR \cite{wang2023exploiting} that employs LDM \cite{rombach2022high}. For a fair comparison, we retrain SR3 on blind settings. We also list the start-of-the-art KP method DCLS \cite{9879171} for reference. As shown in \cref{table:dms}, CDFormer achieves significant improvement in terms of both objective (PSNR and SSIM) and subjective (FID and LPIPS) metrics compared to diffusion-based methods. SR3 and StableSR that adopt pre-trained diffusion models both show expensive costs in time and model size. Our method instead treats the diffusion model as an estimator, therefore CDFormer can address their limitation and is comparable to DCLS.

We next compare our method with state-of-the-art BSR approaches on noise-free degradations with only isotropic Gaussian kernels under blind settings, including the DP methods DASR \cite{wang2021unsupervised} and KDSR \cite{xia2022knowledge}, KP method DCLS \cite{9879171}, and a non-blind SR method RCAN \cite{zhang2018image} for reference. As shown in \cref{table:iso}, the quantitative results indicate that CDFormer can surpass existing BSR methods. Specifically, CDFormer outperforms the state-of-the-art method DCLS by up to $0.71$ dB on Urban100 at a scale of $\times 4$ in all blind settings. However, DCLS relies heavily on accurate kernel estimation, and thus risks significant performance drops in the case of inaccurate estimation or complex degradations. Among DP methods, either DASR or KDSR may lack stability in degradation extraction. In contrast, CDFormer which estimates both content and degradation representations, can reconstruct SR results with sharper and more harmonious textures, evident from the qualitative results in \cref{fig:urban}, demonstrating the efficiency of the CDP in properly guiding  CDFormer and enabling robustness. 
\\
\textbf{General Degradation with Anisotropic Gaussian Kernels and Noise}. To better evaluate our method in the setting of complex degradations, we follow DASR \cite{wang2021unsupervised} that uses $9$ blur kernels and different noise levels. For a fair comparison, we use DnCNN \cite{7839189} as a denoising module to preprocess images for some methods that inherently lack the ability to perform denoising (RCAN \cite{zhang2018image} and DCLS \cite{9879171}). The quantitative results are listed in \cref{table:aniso}. Compared to KDSR, CDFormer achieves a PSNR improvement ranging from $0.1$ dB to $0.5$ dB over all settings. Notice that our method achieves the most significant PSNR improvement when the noise level is set to $0$. However, improvement is limited as the noise level increases. We attribute this phenomenon to the limitations of SR networks, when given severely damaged LR images, having too minimal information to reconstruct ideal SR images. Nevertheless, CDFormer incorporating CDP performs better texture reconstruction in most settings. The qualitative comparison in \cref{fig:aniso} further indicates that CDFormer with the guidance of both content and degradation representations can generate more accurate and clear textures. More results are provided in the supplementary material.

\begin{table}[t]
	\caption{Ablation study of CDP on Set5. Best in \textbf{blod}.}
    \vspace{-6pt}
	\centering
	\label{table:ab of cdp}
	\resizebox{0.47\textwidth}{!}{
		\begin{tabular}{ccccccc}
			\hline
			\multirow{2}{*}{Method} & \multirow{2}{*}{Degradation} & \multirow{2}{*}{Content} &
            \multicolumn{2}{c}{kernel width=1.2} & \multicolumn{2}{c}{kernel width=2.4}  \\
			& & & PSNR $\uparrow$ & SSIM $\uparrow$ & PSNR $\uparrow$ & SSIM $\uparrow$ \\
			\hline
			model1  & \ding{56} & \ding{56} &
			32.074 & 0.8922 & 32.006 & 0.8894   \\
			model2  & \ding{52} & \ding{56} &
			32.372 & 0.8958 & 32.167 & 0.8902   \\
			model3   & \ding{56} & \ding{52} &
			32.451 & 0.8959 & 32.287 & 0.8909   \\
			model4 (Ours)   & \ding{52} & \ding{52} &
			\textbf{32.564} & \textbf{0.8981} & \textbf{32.393} & \textbf{0.8926}  \\
			\hline
	\end{tabular}}
     \vspace{-1em}
\end{table}

\begin{figure}[!htbp]
	\centering
	\begin{subfigure}[b]{0.07\textwidth}
		\centering
		\includegraphics[width=1.0\textwidth]{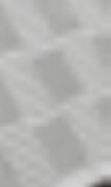}
		\subcaption*{LR}
		\label{lr:92}
	\end{subfigure}
	\begin{subfigure}[b]{0.07\textwidth}
		\centering
		\includegraphics[width=1.0\textwidth]{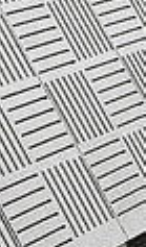}
		\subcaption*{HR}
		\label{hr:92}
	\end{subfigure}
	\begin{subfigure}[b]{0.07\textwidth}
		\centering
		\includegraphics[width=1.0\textwidth]{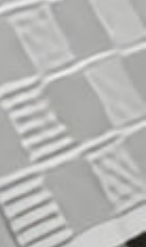}
		\subcaption*{model1}
		\label{model1}
	\end{subfigure}
		\begin{subfigure}[b]{0.07\textwidth}
		\centering
		\includegraphics[width=1.0\textwidth]{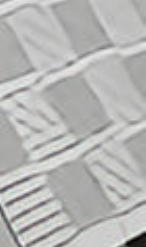}
		\subcaption*{model2}
		\label{model2}
	\end{subfigure}
	\begin{subfigure}[b]{0.07\textwidth}
		\centering
		\includegraphics[width=1.0\textwidth]{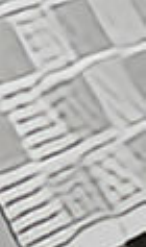}
		\subcaption*{model3}
		\label{model3}
	\end{subfigure}
	\begin{subfigure}[b]{0.07\textwidth}
		\centering
		\includegraphics[width=1.0\textwidth]{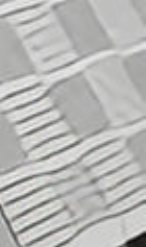}
		\subcaption*{model4}
		\label{model4}
	\end{subfigure}
	\caption{Visualization of ablation study for CDP.}
	\label{alb2}
\end{figure}
\begin{figure}[htbp]
	\centering
    \vspace{-12pt}
	\begin{subfigure}[b]{0.22\textwidth}
		\centering
		\includegraphics[width=1.0\textwidth]{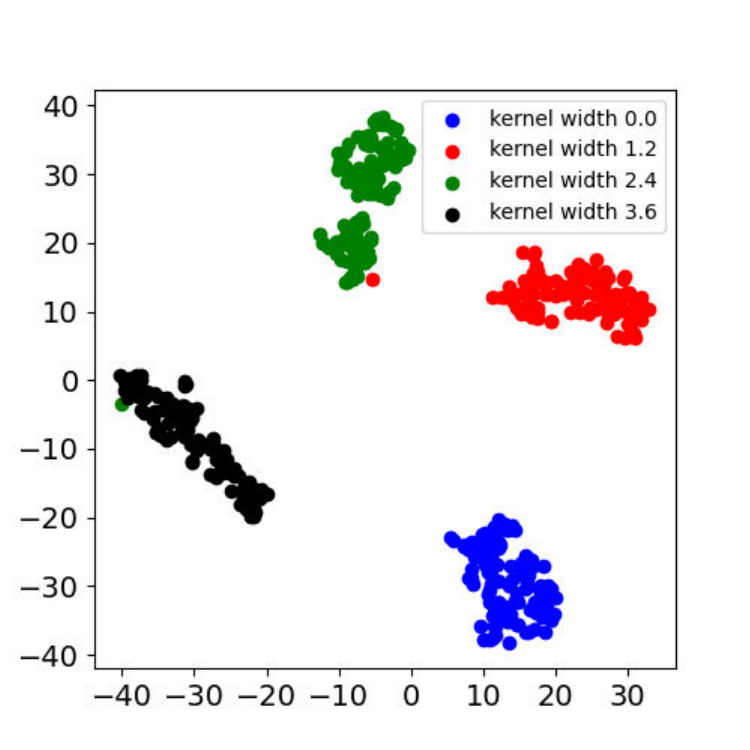}
		\subcaption{CDFormer}
		\label{cdp:degradation}
	\end{subfigure}
	\begin{subfigure}[b]{0.22\textwidth}
		\centering
		\includegraphics[width=1.0\textwidth]{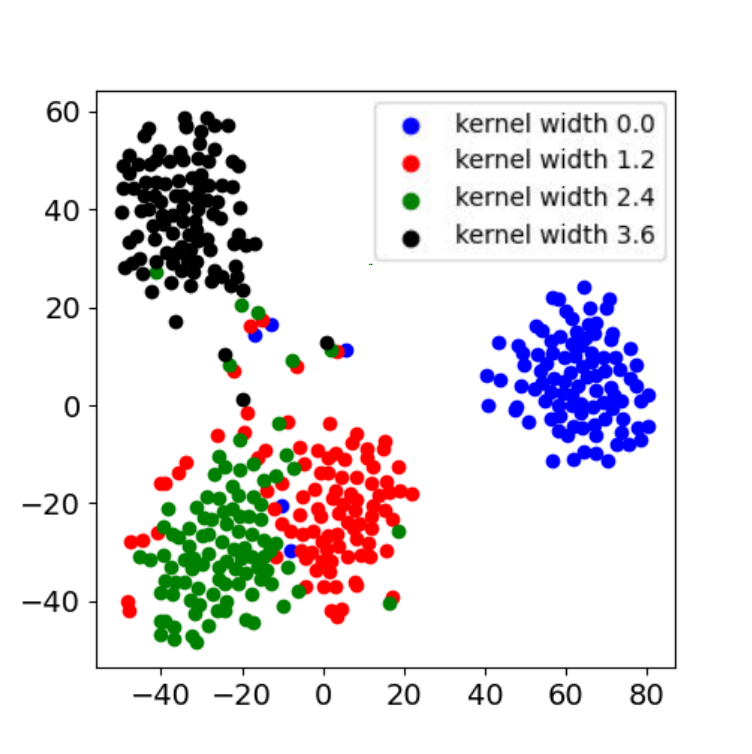}
		\subcaption{DASR\cite{wang2021unsupervised}}
		\label{cdp:dasr}
	\end{subfigure}
	\caption{T-SNE results of CDFormer and DASR.}
    \vspace{-10pt}
	\label{fig:cdp}
\end{figure}

\subsection{Ablation Study}
In this section, we will investigate several key components of CDFormer. All the following experiments are conducted on Set5 \cite{bevilacqua2012low} dataset with scale factor $4$. We also provide ablation experiments for iterations number $T$ and CDRB in the supplementary material.
\\

\textbf{Effects of Content Degradation Prior (CDP)}. 
We first perform an ablation study on the proposed CDP  under the lightweight settings. As listed in \cref{table:ab of cdp}, compared to model1 with no prior information, model2 with degradation prior, and model3 with content prior, our method (model4) with the guidance of both content and degradation representations can achieve the best performance.
\cref{alb2} demonstrates the improvement of CDP visually, further proving the effectiveness of CDP. Our baseline model is able to reconstruct SR images with more accurate textures and fewer distortions, highlighting the effects of CDP on preserving image content and regulating structural textures.

\begin{figure}[htbp]
	\centering
	\begin{subfigure}[b]{0.23\textwidth}
		\centering
		\includegraphics[width=1.0\textwidth]{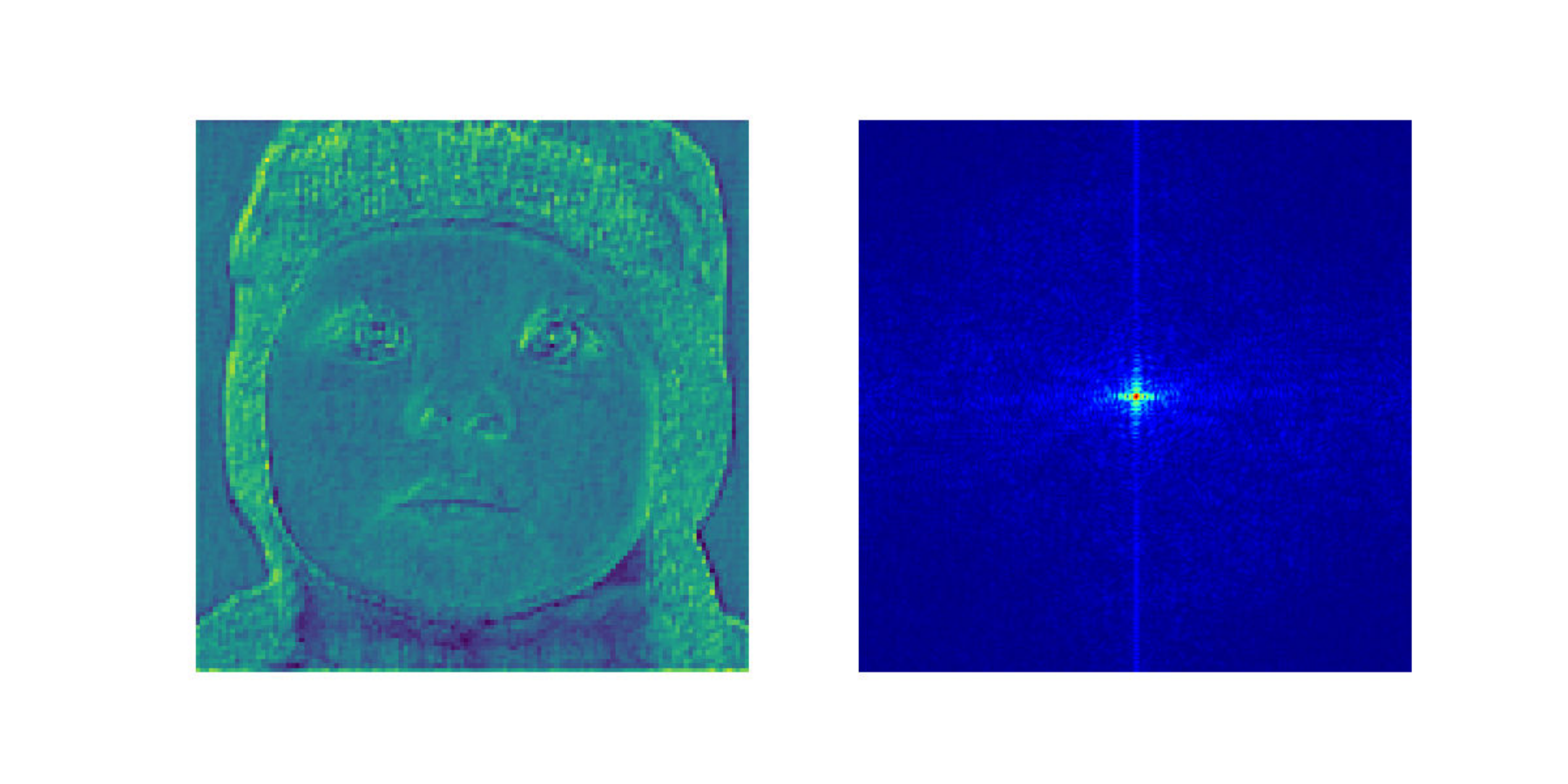}
		\subcaption{w/o CDP}
		\label{cdp:wo_cdp}
	\end{subfigure}
	\begin{subfigure}[b]{0.23\textwidth}
		\centering
\includegraphics[width=1.0\textwidth]{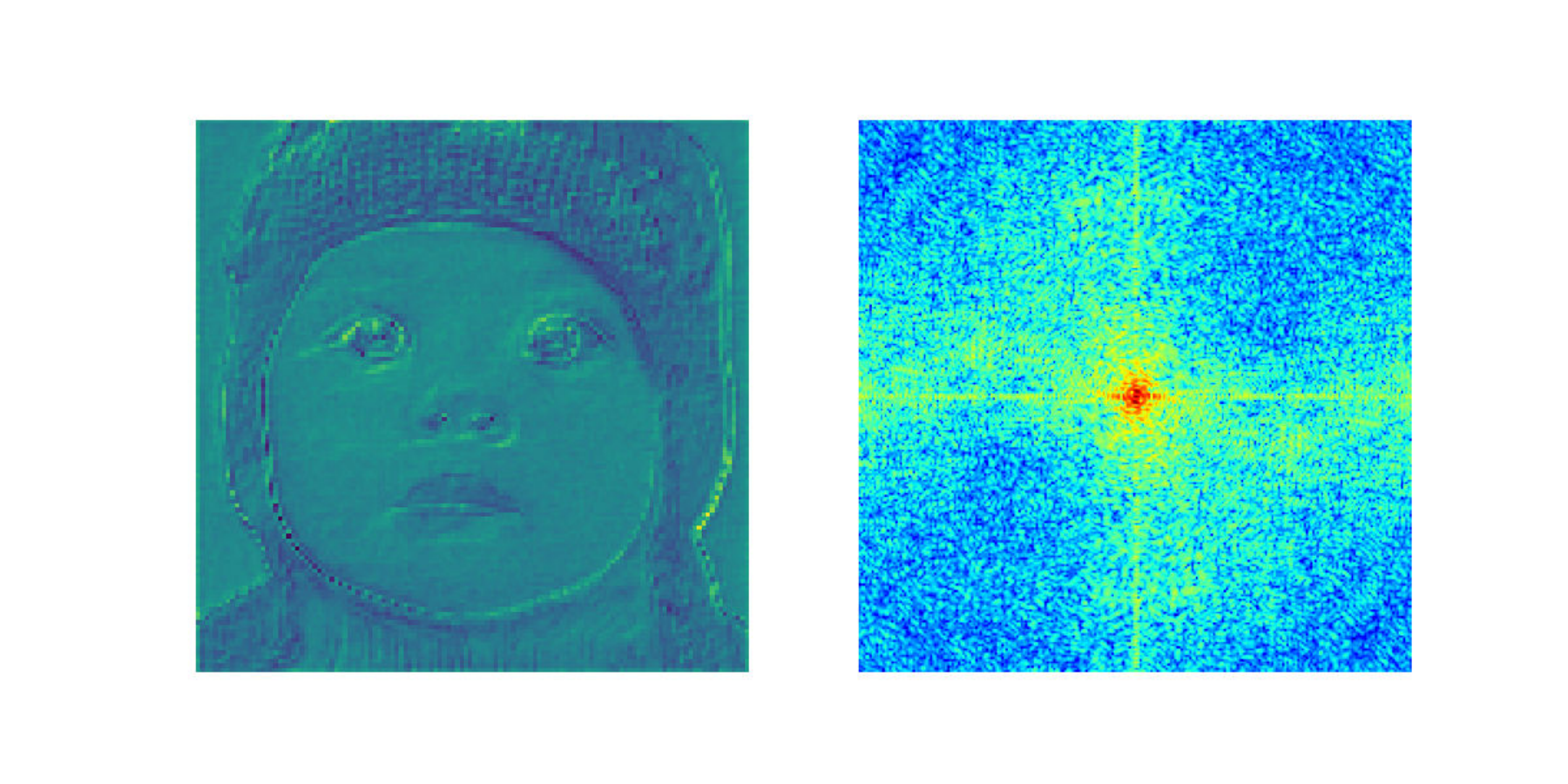}
		\subcaption{w/ CDP}
		\label{cdp:w_cdp}
	\end{subfigure}
	\caption{Visualization for feature maps and Fourier plots of CDP.}
    \vspace{-15pt}
	\label{fig:wo/w CDP}
\end{figure}

To compare the learned representations between previous DP methods and our proposed CDFormer, we utilize T-SNE \cite{van2008visualizing} to visualize representations in \cref{fig:cdp}. Specifically, we use $100$ images with $4$ different degradation kernels to generate LR images. The visualization comparison indicates that CDFormer exhibits superior clustering results. It is notable that DASR employs the contrastive learning mechanism to push away the different degradations and pull close the same degradations. In contrast, CDFormer without any explicit estimation is able to distinguish different degradations, which is also verified by LAM \cite{9578694} visualization provided in the supplementary material. Furthermore, we utilize the Fourier transform to visualize and analyze the effect of CDP. The results in \cref{fig:wo/w CDP} explain that CDP can strengthen low-frequency components. This further confirms that features with CDP contain a greater abundance of low-frequency information, which can be assumed to be local content details, resulting in reconstructed images with more harmonious textures and clear edges.


\textbf{Effects of Diffusion Model.} We next conduct an ablation study on the diffusion model. We employ a two-stage strategy that $E_{GT}$ trained in the first stage will supervise the training of $E_{LR}$ and the diffusion model in the second stage. Notice that the content details in $E_{GT}$ are extracted from HR images, while CDFormer in the second stage utilizes the LR images only and recreates CDP via the diffusion process. As demonstrated in \cref{table:ab of dm}, diffusion model with the conditional vector from LR images is capable of generating reasonable content and degradation representations, even surpassing the ground truth model. This observation indicates the powerful ability of the diffusion model to capture the data distribution and also confirms the superiority of our method.
\begin{table}[t]
	\caption{Ablation study of diffusion model on Urban100.}
    \vspace{-6pt}
	\centering
	\label{table:ab of dm}
	\resizebox{0.46\textwidth}{!}{
		\begin{tabular}{ccccccc}
			\hline
			\multirow{2}{*}{Method} & 
            \multicolumn{2}{c}{kernel width=0} & \multicolumn{2}{c}{kernel width=1.2} \\
			& PSNR $\uparrow$ & SSIM $\uparrow$ & PSNR $\uparrow$ & SSIM $\uparrow$ \\
			\hline
			CDFormer (Stage1)   &
			26.945 & 0.8121 & 26.923 & 0.8096 \\
			CDFormer (Stage2)  &
			\textbf{27.210} & \textbf{0.8189} & \textbf{27.058} & \textbf{0.8116}\\
			\hline
	\end{tabular}}
    \vspace{-1em}
\end{table}

%% file: sec/5_conclusion.tex
\section{Conclusion}
In this paper, we propose a novel Blind image Super-Resolution method dubbed CDFormer. The distinctiveness of our proposed method lies in the idea of introducing the concept of Content Degradation Prior (CDP) to provide rich low- and high-frequency information for reconstruction. The diffusion model is cleverly used as an estimator to recreate the CDP from the LR images. Since the vector to be learned is one-dimensional, our denoising process is more efficient in sampling and computation. We redesign an adaptive SR network to take full advantage of CDP via injection modules as well as the interflow mechanism to enhance feature representation. Various experiments show that CDFormer not only achieves improved performance with more accurate texture reconstruction in complex degradation scenarios, but also confirms the diffusion model with infinite possibilities in super-resolution.
\vspace{-5pt}
\section*{Acknowledgement}
This work was partially supported by the National Natural Science Foundation of China (No. 62272227 \& No. 62276129), and the Natural Science Foundation of Jiangsu Province (No. BK20220890).
